\documentclass[journal]{IEEEtran}

\usepackage{cite}
\usepackage{algorithm}
\usepackage{algorithmic}
\usepackage{graphicx}
\usepackage{tabularx}
\usepackage{subcaption}
\usepackage{url}
\usepackage{refcount}
\usepackage{amsmath}
\usepackage{subcaption}
\usepackage{makecell}
\usepackage{xcolor}

\begin{document}

\title{LLM-Powered Swarms: \\A New Frontier or a Conceptual Stretch?}

\author{Muhammad~Atta~Ur~Rahman,~Melanie~Schranz,~Samira~Hayat\\
Lakeside Labs GmbH \\

Klagenfurt, Austria\\
\{rahman,schranz,hayat\}@lakeside-labs.com

}

\maketitle

\begin{abstract}
Swarm intelligence describes how simple, decentralized agents can collectively produce complex behaviors. Recently, the concept of swarming has been extended to large language model (LLM)-powered systems, such as OpenAI’s Swarm (OAS) framework, where agents coordinate through natural language prompts. This paper evaluates whether such systems capture the fundamental principles of classical swarm intelligence—decentralization, simplicity, emergence, and scalability. Using OAS, we implement and compare classical and LLM-based versions of two well-established swarm algorithms: Boids and Ant Colony Optimization. Results indicate that while LLM-powered swarms can emulate swarm-like dynamics, they are constrained by substantial computational overhead. For instance, our LLM-based Boids simulation required roughly 300$\times$ more computation time than its classical counterpart, highlighting current limitations in applying LLM-driven swarms to real-time systems.
\end{abstract}


\IEEEpeerreviewmaketitle

\section{Introduction}

\IEEEPARstart{S}{warm} intelligence continues to attract significant attention from researchers and engineers. In nature, swarming systems exist as flocks of birds, schools of fish, and colonies of ants, where they are characterized by local interactions among agents following simple rules. These interactions give rise to global patterns and adaptive behaviors that are greater than the sum of their parts~\cite{schranz2021swarm}. However, the term ``swarm" has recently been appropriated in novel contexts, such as OpenAI's Swarm (OAS) framework~\cite{openai_swarm}, where the dynamics and mechanisms differ from their traditional counterparts. This paper explores the differences, examining how the principles that define classical swarm algorithms translate, or fail to translate, within large language model (LLM)-based systems such as OAS, which is selected as a representative framework for LLM-powered swarms in this paper.

At the heart of traditional swarms is decentralization~\cite{Bonabeau1999, schranz2021swarm}: each agent operates independently, responding only to its immediate environment and neighbors, yet collectively achieving robust and scalable global behavior. In our LLM-based implementations, swarm behavior is distributed across multiple specialized agents, each guided by a prompt responsible for a specific behavioral rule. It is important to emphasize that these ``agents” differ fundamentally from classical swarm agents---they are prompt-driven elements powered by LLM rather than autonomous entities. This design preserves decentralization while embedding language-driven reasoning within individual components, enabling a direct comparison with classical swarm algorithms.

This work selects two well-established swarm algorithms—Boids and Ant Colony Optimization (ACO)—to evaluate how their classical implementations compare with LLM-based counterparts. OAS can leverage both local and cloud-based LLMs for implementation, each of which presents distinct advantages and trade-offs in computational cost, accessibility, and scalability. We evaluate both configurations to select a representative model for comparison based on performance. 

This comparison not only investigates whether LLM-powered swarms embody the core principles of swarm intelligence, but also highlights their potential---and current limitations. One notable observation was that current LLM-based implementations exhibit significant computational overhead, making real-time deployment in practical swarm systems challenging.

The paper is structured as follows:  In Sec.~\ref{sec:Background}, we provide an overview of the OAS framework and its possible implementations: local versus cloud-based. We then provide the system description in Sec.~\ref{sec:SystemDescription}, LLM implementation evaluation in Sec.~\ref{sec:LLMEvaluation}, and a comparison of classical and LLM-based implementations of Boids and ACO in Sec.~\ref{sec:Evaluation}. We conclude with a summary of key findings in Sec.~\ref{sec:Conclusions}.

\section{Background}
\label{sec:Background}
Advances in LLMs have introduced a new paradigm for multi-agent coordination, motivating a reevaluation of how swarm intelligence can be implemented. Frameworks such as OAS provide a structured means of constructing agent-based systems that use LLMs for coordination and task execution. Understanding how these LLM-driven swarms align with or diverge from classical, rule-based swarm systems is essential for assessing their potential as decentralized intelligent systems.


\subsection{OpenAI's Swarm Framework}

The OAS framework provides a platform for creating multi-agent systems driven by language models. Each agent can be assigned specialized tasks, receiving prompts that are processed and parsed to generate actions, enabling modular and decentralized behavior. OAS supports both cloud-based LLMs, such as GPT-4o, and local models like Llama~\cite{touvron2023llamaopenefficientfoundation} or Qwen~\cite{bai2023qwentechnicalreport}, allowing flexible deployment while maintaining agent-level coordination for complex workflows.

\subsection{Local versus Cloud-based LLMs}

Using the OAS framework requires either a locally deployed or cloud-hosted LLM model. The choice between the two affects factors such as computational requirements and performance.

\subsubsection{Local LLMs}

A wide range of local LLMs is now available, ranging from lightweight 1B-parameter models to large-scale architectures that exceed 90B parameters~\cite{huggingface2025model}. In practice, model size directly affects both performance and hardware requirements: larger models demonstrate stronger reasoning and generalization capabilities, but they also require significantly more GPU memory to operate efficiently.

To understand these requirements, it is essential to relate the number of model parameters to the memory needed for deployment. The effective size of an LLM depends not only on its parameter count \( N \), but also on its numerical precision \( b \), such as FP32, FP16, or quantized formats (Q8, Q4)~\cite{gholami2021survey}. The corresponding memory demand \( M \) (in GB) can be approximated by:

\begin{equation}
\label{eq:memory}
M = \frac{N \times b}{8 \times 1024^3}
\end{equation}

This relationship highlights the trade-off between precision and resource usage. For instance, a 1B-parameter model requires roughly 3.7~GB in FP32 precision, but only about 0.47~GB when quantized to Q4. Despite this reduction through quantization, large models pose large memory requirements. Using Equation~\ref{eq:memory}, the VRAM requirements for Llama~90B and Qwen~32B with Q4 quantization approximate to 40~GB and 15~GB, respectively. As of October~2025, even high-end consumer GPUs like NVIDIA’s GeForce RTX~5090 (32~GB VRAM)~\cite{nvidia_geforce_rtx_5090} cannot accommodate such models fully, necessitating data center GPUs like the A100~\cite{9361255} or H100~\cite{10070122}, which provide higher memory capacities.

Popular local models include Meta’s Llama~\cite{touvron2023llamaopenefficientfoundation}, Alibaba’s Qwen~\cite{bai2023qwentechnicalreport}, Google’s Gemma~\cite{gemmateam2024gemma}, and Mistral AI’s Mistral~\cite{jiang2023mistral7b}. In this work, Llama and Qwen are used as representative examples due to their accessibility and compatibility with OAS.

\paragraph{Meta's Llama}  

Llama is a family of research-focused LLMs for diverse natural language processing tasks~\cite{touvron2023llamaopenefficientfoundation}. Since its debut, it has evolved through multiple versions, with Llama 3.2 introducing multimodal support for both text and images. Llama models range from 1B to 90B parameters. This offers a wide choice to address the trade-off between performance and hardware requirements. For example, a 1B-parameter model corresponds to approximately 1.86~GB in FP16 precision, while a 90B-parameter model requires roughly 167~GB.

\paragraph{Alibaba's Qwen}  

Qwen is optimized for commercial use, with strong support for many languages including Chinese and English~\cite{bai2023qwentechnicalreport}. Released in 2023, Qwen has progressed through multiple versions, with Qwen 3 introducing multimodal capabilities and model sizes ranging from 0.6B to 235B parameters~\cite{yang2025qwen3}.

\subsubsection{Cloud-Based GPT}

Cloud-hosted models like OpenAI's GPT-4 offer scalable access via APIs. Since GPT-1~\cite{radford2018improving} in 2018, the series has advanced considerably, with GPT-5~\cite{wang2025capabilitiesgpt5multimodalmedical} (2025) introducing major improvements in reasoning, adaptability, and multimodal capabilities. 

Choosing between local and cloud-hosted language models entails trade-offs in accessibility, cost, and computational requirements, as shown in Table~\ref{globallocal_llm}. While cloud-based options such as OpenAI’s GPT, Anthropic’s Claude, and DeepSeek~\cite{ArtificialAnalysis2025} offer convenience and high performance, they typically involve subscription fees and paid API usage. In contrast, local LLMs can be deployed freely on personal hardware, providing unrestricted experimentation within OAS without recurring costs, although they demand substantial computational resources upfront.

\begin{table}[h]
\caption{Cloud-based vs. local LLMs}
\label{globallocal_llm}
\centering
\small
\setlength{\tabcolsep}{4pt}
\resizebox{1\linewidth}{!}{
\begin{tabular}{|p{1.6cm}|p{3.5cm}|p{3cm}|}
\hline
\textbf{Aspect} & \textbf{Cloud-based} & \textbf{Local} \\
\hline
Performance & Handles large, complex tasks & Limited by hardware \\
\hline
Accessibility & Multi-device access & Single machine only \\
\hline
Cost & API fees & Upfront hardware cost, no fees \\
\hline
Privacy & Data processed remotely & Full local control \\
\hline
Latency & Dependent on communication network& Instant, offline use \\
\hline
\end{tabular}
}
\end{table}

\section{System description}
\label{sec:SystemDescription}
To explore the differences between classical and LLM-based swarm approaches, example swarm algorithms were implemented using both methods. This section outlines the system architecture, including hardware and software configurations, the language models employed, and the implementation details of each approach.

\subsection{Hardware and Software}

The test system featured an AMD Ryzen 9 8545HS processor (8 cores, 16 threads), 32~GB of system memory, and an NVIDIA GeForce RTX~4070 GPU with 8~GB of VRAM---the most critical hardware component. 

LM Studio~\cite{lmstudio} was used to load the models, and OAS was adapted to redirect API calls to LM Studio and access local LLMs instead of GPT.

\subsection{Utilized LLMs}

Because of the limited 8GB of available VRAM, models exceeding 7 billion parameters perform poorly unless heavily quantized. While memory offloading to system RAM is possible, it greatly reduces computational speed.

The local models tested include Llama 1B, 3B, 7B, and a quantized 14B, along with similarly sized Qwen models. Next, GPT API was utilized to access the latest cost-effective models, allowing a direct comparison with the locally deployed LLMs.

\subsection{Swarm Algorithms}
\label{sec:implementations}

We used Boids and ACO to compare classic and LLM-driven swarm implementations, as they clearly demonstrate how swarm systems work. Boids is a rule-based algorithm that simulates flocking behavior, while ACO is an ant-inspired heuristic to find optimal paths.

\paragraph{The Boids Model}

The Boids model simulates emergent flocking behavior using three simple rules that each agent (or ``boid'') follows,

\begin{enumerate}
    \item \textbf{Separation:} Each boid steers away from nearby neighbors to avoid overcrowding. This is done by checking the positions of surrounding boids within a certain radius and applying a repulsive force.
    
    \item \textbf{Cohesion:} Each boid moves towards the average position of its local neighbors. This encourages group cohesion and helps maintain the structure of the flock.
    
    \item \textbf{Alignment:} Each boid adjusts its velocity to match the average velocity of its nearby neighbors, resulting in coordinated and aligned movement within the group.
\end{enumerate}

For the classical rule-based Boids model, a Python simulation was developed, where each boid independently applied these three rules at each time step to update its position and velocity.

For the LLM-powered alternative, the same behavioral logic was replicated using LLM prompts, where each boid is treated as an independent reasoning unit. Each boid issues three separate prompts, one for each rule, to update its state based on nearby boids. This setup aims to simulate a fully decentralized system, similar to the traditional implementation, but powered by natural language reasoning instead of hard-coded logic. Importantly, in both the rule-based and LLM-based implementations, the design is modular: each behavior (separation, cohesion, and alignment) is defined and executed independently, and the overall flocking dynamics emerge from their composition. This modular structure ensures that the effects of individual behaviors can be analyzed and debugged.

\paragraph{The ACO Model}

ACO is inspired by the emergent foraging behavior of real ants, particularly how they find the shortest paths between their nest and food sources. The ants iteratively discover optimal solutions through indirect communication via pheromones using three mathematical rules:

\begin{enumerate}
    \item \textbf{Path Selection:} Each ant chooses a path based on a probabilistic decision rule that considers the amount of pheromones on each possible path and the distance to the destination. Paths with higher pheromone concentration and shorter distances are more likely to be selected.
    
    \item \textbf{Pheromone Update:} After completing a path, the ant deposits pheromones on the path it traveled. This reinforces the chosen path, making it more attractive to other ants in future iterations.
    
    \item \textbf{Pheromone Evaporation:} Over time, pheromone levels in the environment decrease due to evaporation. This prevents the algorithm from converging too early on suboptimal paths.
\end{enumerate}

Following the same approach as in the Boids' model, a baseline classical implementation was developed using hard-coded rules in Python. In parallel, an LLM-powered version was constructed where the three rules were implemented using separate LLM prompts at each iteration. 

It is important to note here that an LLM agent does not correspond to an algorithm agent, i.e., boid and ant. An LLM agent is used here to implement the algorithm rules; for ACO, the path selection and pheromone update (LLM) agents are rules for the ant (swarm) agent while the pheromone evaporation (LLM) agent is a rule for the environment.

For the interested reader, our GitHub repository~\cite{atta2026swarms} contains the code and parameters for reproducibility, system diagrams and additional plots.

\section{LLM Platform Evaluation}
\label{sec:LLMEvaluation}
To evaluate how LLM-based swarms compare to traditional rule-based swarms, we first identified a representative LLM model for the experiments. This involved comparing locally hosted models with cloud-based GPT to determine the model offering the best performance. A key part of this process was developing an effective prompt design strategy to ensure consistent and reliable responses across models. The models were evaluated for response latency and system resource utilization. The model demonstrating the best overall performance was selected for the final comparison against classic swarm implementations.

\subsection{Prompt Design Strategy}

Prompt engineering was essential for obtaining reliable results. We used an iterative process in which we used GPT to generate candidate prompts from explicit swarm-rule requirements, such as separation, cohesion, or pheromone update. These prompts were then tested and refined, with particular attention to strict output formatting to prevent parsing and execution failures.

The process consisted of four steps:
\begin{enumerate}
\item \textbf{Specification:} Define the required rule behavior and output constraints.
\item \textbf{Generation:} Let the LLM generate a candidate prompt.
\item \textbf{Evaluation:} Execute the prompt and assess the output for correctness and formatting.
\item \textbf{Refinement:} Correct identified issues and repeat steps 3 and 4 until the output is stable and usable.
\end{enumerate}

This approach allows LLMs to contribute to the design of their own prompts, while human input focuses on defining constraints and evaluating reliability. Its effectiveness therefore depends on the clarity of the initial specification and the quality of human guidance.

\subsection{LLM Prompts}
\label{appendix:boids_prompts}

The LLM prompts used to implement the three boid behaviors are:
\begin{itemize}
    \item \textbf{Separation Behavior Agent}: You are a boid at position \{position\} with current velocity \{velocity\}. 
Other boids: \{other\_boids\}. Your task is to avoid getting too close to 
other boids within a radius of \{radius\}. Return a (dx, dy) vector 
representing the separation force to apply to your velocity. Only output the vector as 
(dx, dy). No additional text.
    \item \textbf{Cohesion Behavior Agent:} You are a boid at position \{position\} with current velocity \{velocity\}. 
Other boids: \{other\_boids\}. Your task is to move slightly toward the 
average position of nearby boids within a radius of \{perception\_radius\}. 
Return a (dx, dy) vector representing the cohesion force to apply to your 
velocity. Only output the vector as 
(dx, dy). No additional text.
    \item \textbf{Alignment Behavior Agent:} You are a boid at position \{position\} with velocity \{velocity\}. Other 
boids: \{other\_boids\}. Your task is to align your velocity with the average 
velocity of nearby boids within a radius of \{perception\_radius\}. Return a 
(dx, dy) vector representing the alignment force to apply to your velocity. Only output the vector as (dx, dy). 
No additional text.
\end{itemize}
\hfill \break
The prompts for LLM ACO are:

\begin{itemize}
\item \textbf{Path Selection Agent}: You are an ant in an ACO simulation using an exploration---exploitation strategy. Current step: \{step\} of \{max\_iterations\}. Paths: \{paths\}. Strategy: Early Phase (steps 0---\{early\_phase\_end\}): Explore---choose paths more randomly, aiming for roughly 50/50 exploration. Middle Phase (steps \{mid\_phase\_end\} to \{late\_phase\_start\}): Transition---start considering pheromones but continue occasional exploration. Late Phase (steps \{late\_phase\_start\}): Exploit---focus on the path with the better pheromone-to-distance ratio. Current phase: \{current\_phase\}. In the \{current\_phase\} phase, you should \{phase\_instruction\}. Return only: short or long.
\item \textbf{Pheromone Update Agent:} You are a pheromone update agent in an Ant Colony Optimization simulation. 
The paths are given by \{paths\}. Based on the chosen path which is 
\{chosen\_path\}, update the pheromone levels to reflect the quality of the 
chosen path and respond with the updated pheromones as an output: [x,y] 
where x corresponds to short and y corresponds to long. No additional text.
\item \textbf{Evaporation Agent:} You are a pheromone evaporation agent. The paths are given by \{paths\}. 
Based on the evaporation rate \{evaporation\_rate\}, apply evaporation to 
the pheromone levels and return the pheromone levels in the format: [x,y] 
where x corresponds to short and y corresponds to long. No additional text.
\end{itemize}

\subsection{Local LLMs vs GPT}

\begin{figure*}[ht]
    \centering
    \begin{subfigure}[t]{0.48\textwidth}
        \centering
        \includegraphics[width=\textwidth]{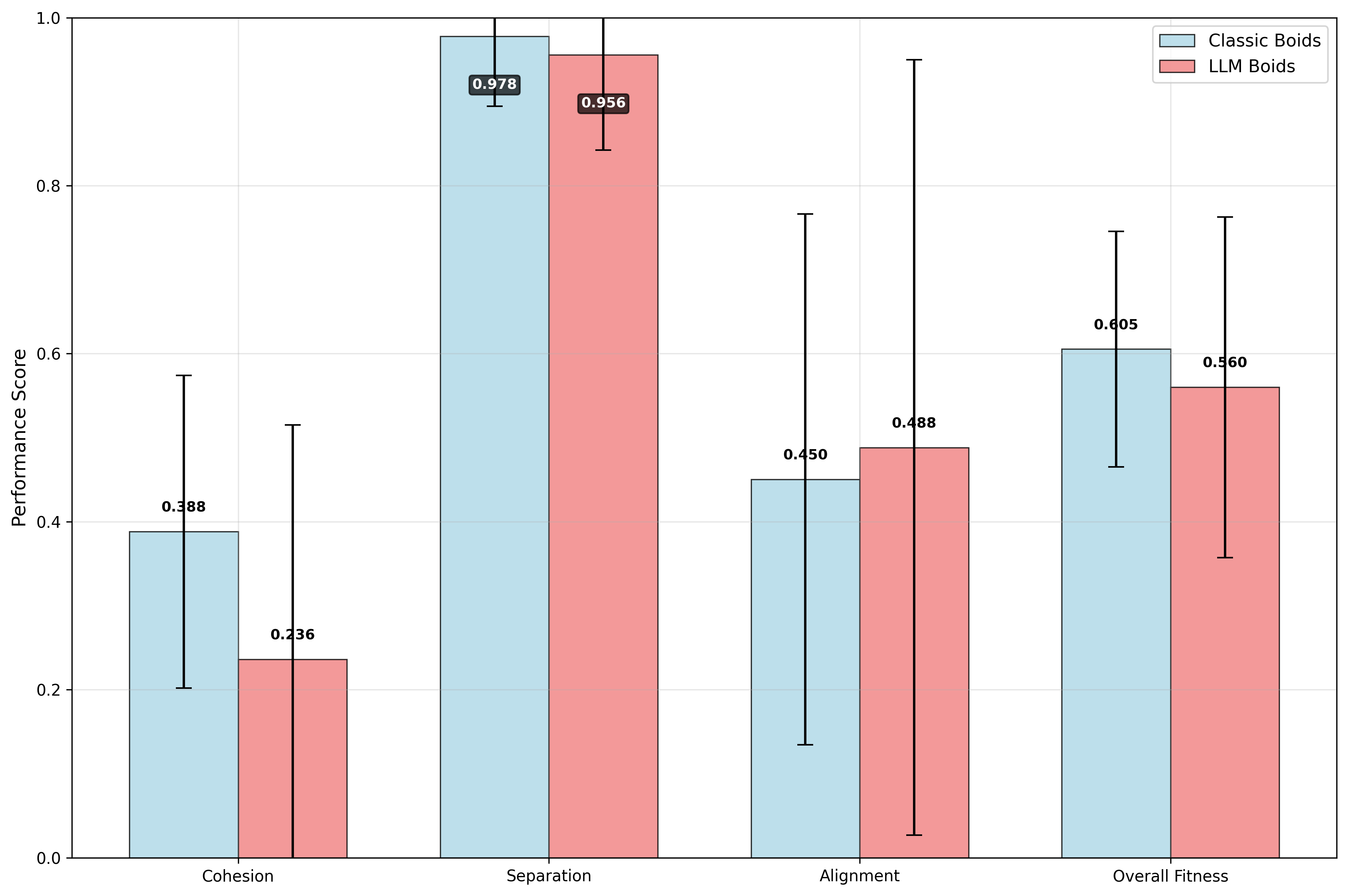}
        \caption{Mean performance metrics with standard deviations.}
        \label{fig:boids_classic}
    \end{subfigure}
    \hfill
    \begin{subfigure}[t]{0.48\textwidth}
        \centering
        \includegraphics[width=\textwidth]{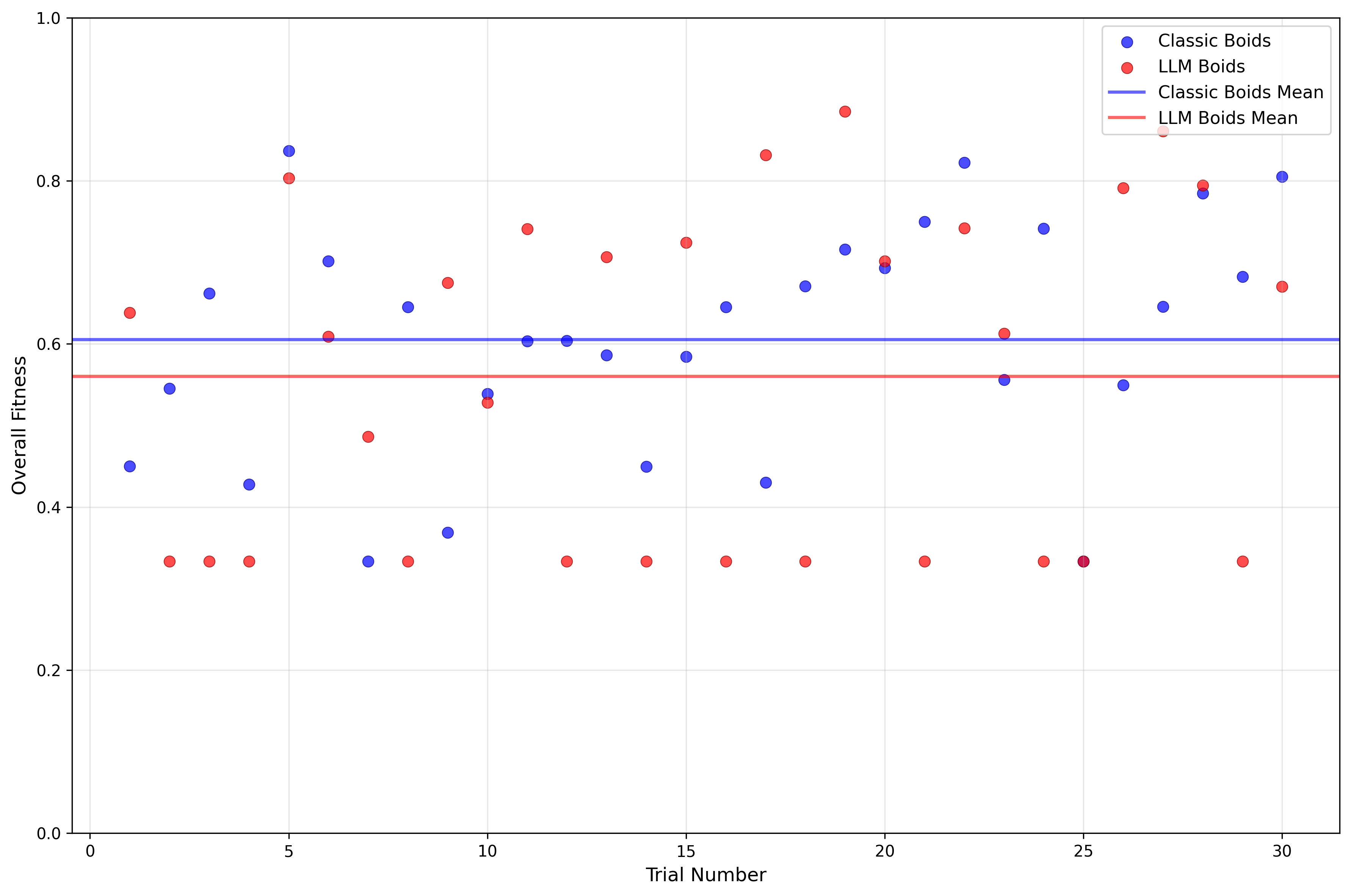}
        \caption{Overall fitness (F) by trial with mean performance indicators.}
        \label{fig:boids_llm}
    \end{subfigure}
    
    \caption{Classic Boids vs LLM Boids performance analysis.}
    \label{fig:boids_classic_llm}
\end{figure*}

During preliminary model selection, smaller local models occasionally produced incorrect numerical outputs, which affected their suitability for the purposes of this work. Larger models showed improved reliability but incurred higher computational costs. We therefore selected Qwen 2.5 Instruct 14B as the representative local model based on the trade-off between output reliability and computational requirements. These preliminary arithmetic observations are used only for selecting the candidate LLM for this work and a detailed arithmetic evaluation is outside the scope of this paper.

We selected the \textbf{Path Selection} rule from the ACO for performance evaluation. For simplicity, only two path options were used: short and long. In language models, \textit{token length} refers to the number of text units—such as words or word fragments—that the model processes, with longer prompts typically resulting in higher latency. Using Qwen’s tokenization scheme, three prompts with varying lengths (approximately 75, 330, and 582 tokens) were created using the prompt design strategy described above. Although different in verbosity and structure, each prompt produced the same output (either short or long). This allowed us to explore the relationship between prompt length and response time.

We evaluated resource utilization and latency for both the local LLM (Qwen 2.5 Instruct, 14B) and the cloud-based GPT model.
Table~\ref{tab:llm_resource_latency} summarizes the results.

\begin{table}[ht]
\caption{Comparison of local (Qwen 2.5 Instruct 14B) and cloud-based (GPT) models \small{(all values are approximate)}}
\label{tab:llm_resource_latency}
\centering
\small
\setlength{\tabcolsep}{4pt}
\resizebox{1\linewidth}{!}{
\begin{tabular}{|p{2.8cm}|>{\centering\arraybackslash}p{3.3cm}|>{\centering\arraybackslash}p{3.1cm}|}
\hline
\textbf{Metric} & 
\makecell{\textbf{Local}\\ \textit{(Qwen 2.5 Instruct 14B)}} & 
\makecell{\textbf{Cloud-based}\\ \textit{(GPT)}} \\
\hline
CPU Usage & 17\% & 7\% \\
\hline
RAM Usage & 77\% & 48\% \\
\hline
GPU Usage & 81\% & 0\% \\
\hline
Latency (75 tokens) & 0.85 s & 0.60 s \\
\hline
Latency (330 tokens) & 1.1 s & 0.9 s \\
\hline
Latency (582 tokens) & 1.8 s & 1.4 s \\
\hline
\end{tabular}
}
\end{table}

As shown, GPT demonstrated significantly lower local resource consumption and faster response times. The elimination of GPU usage and the reduced latency across all prompt lengths established GPT as the candidate of choice for  evaluation of LLM-driven swarms within the scope of this study. For interested readers, additional comparative results between classic and local LLM implementations can be found on our public Github repository~\cite{atta2026swarms}.

\section{Evaluation}
\label{sec:Evaluation}
We implemented Boids and ACO to compare traditional rule-based swarms with LLM-driven swarms. The source code is publicly available~\cite{atta2026swarms}. Each implementation was evaluated to examine trade-offs in performance and behavioral results.

\subsection{Case Study: The Boids Model}

\begin{figure*}[!t]
    \centering
    
    \begin{subfigure}[t]{0.48\textwidth}
        \centering
        \includegraphics[width=\textwidth]{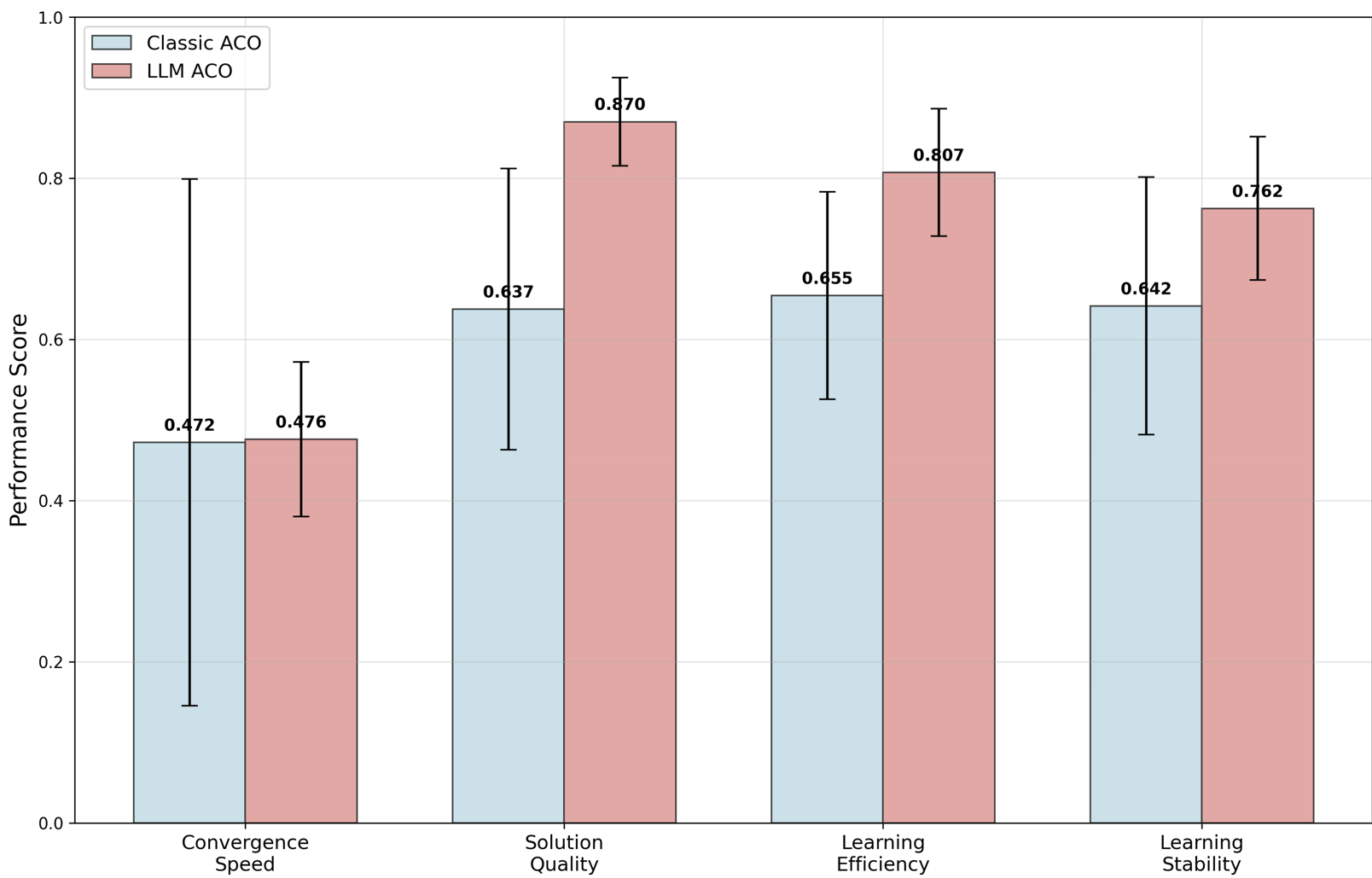}
        \caption{Mean performance metrics with standard deviations.}
        \label{fig:aco_performance}
    \end{subfigure}
    \hfill
    \begin{subfigure}[t]{0.463\textwidth}
        \centering
        \includegraphics[width=\textwidth]{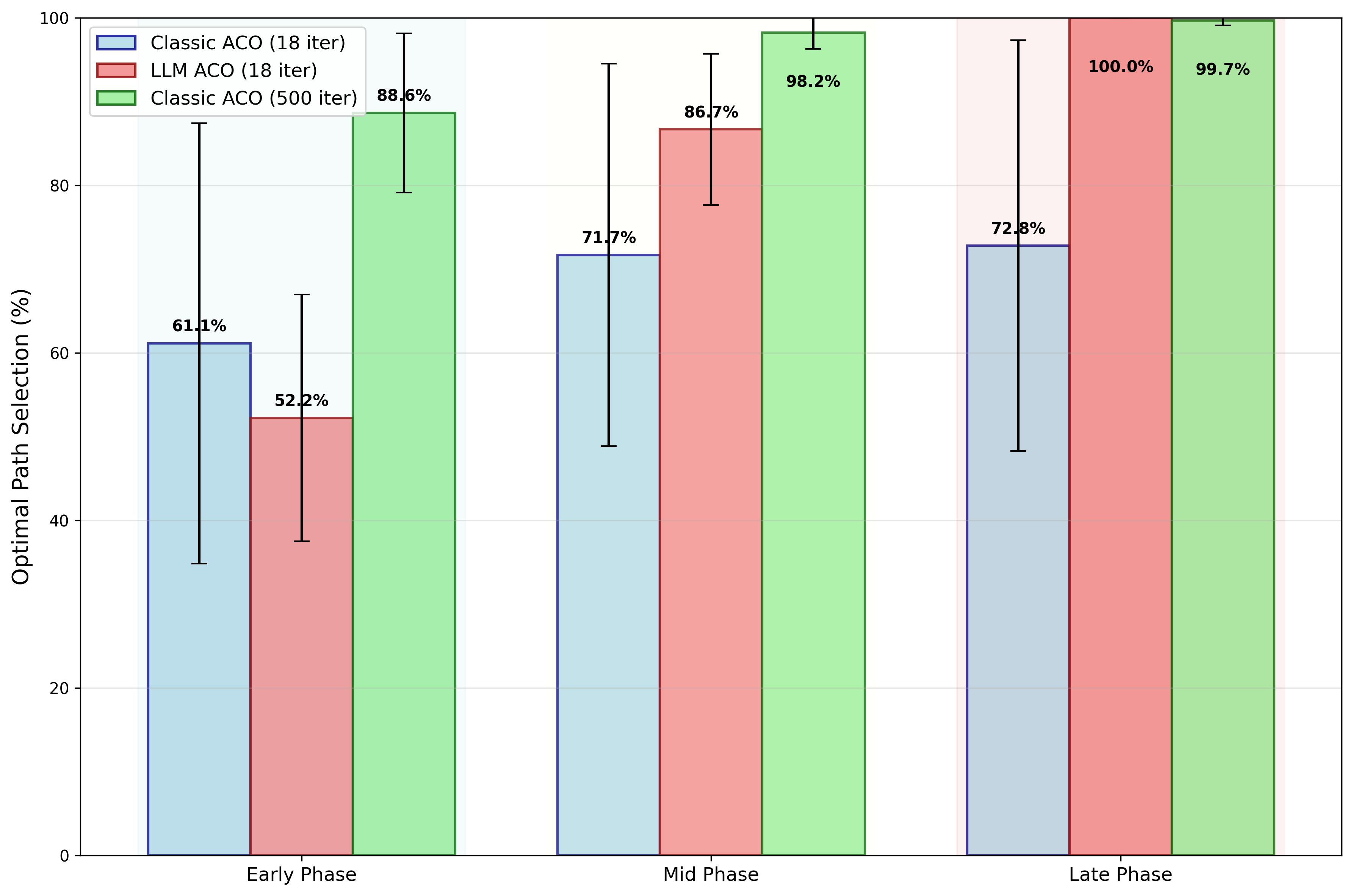}
        \caption{Phase-based optimal path selection.}
        \label{fig:aco_phase}
    \end{subfigure}
    \hfill
    \begin{subfigure}[t]{0.48\textwidth}
        \centering
        \includegraphics[width=\textwidth]{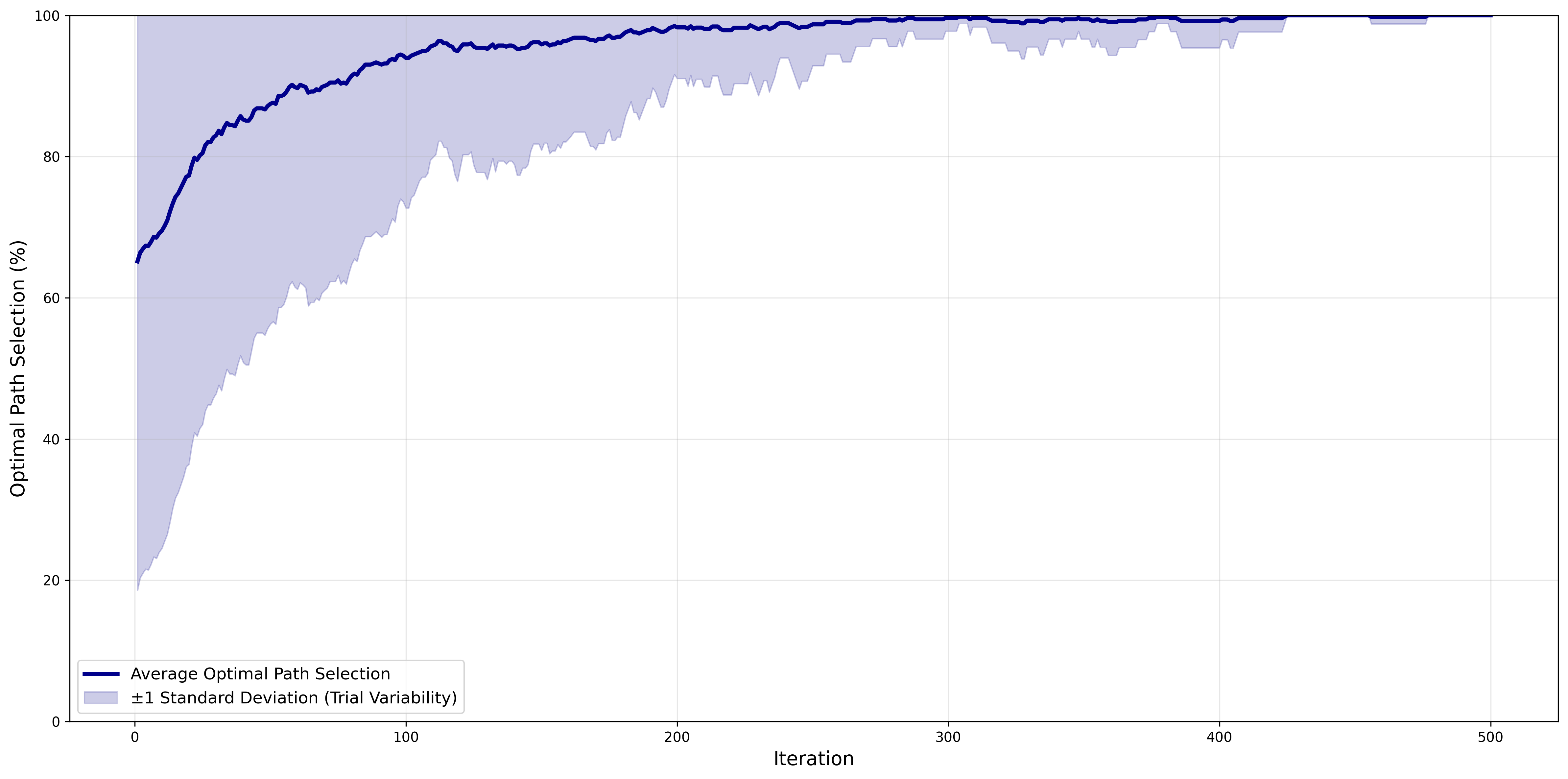}
        \caption{Learning progression over iterations for Classic ACO.}
        \label{fig:aco_multirun}
    \end{subfigure}

    \caption{Comprehensive ACO performance analysis.}
    \label{fig:aco_classic_llm}
\end{figure*}

As previously mentioned, each boid was responsible for executing three core behaviors: separation, cohesion, and alignment. For the LLM-based implementation, this translated into issuing \textit{three separate prompts per boid per time step}~(see Section~\ref{appendix:boids_prompts}). The output vectors from each prompt were parsed and combined to update each boid’s velocity.

To ensure a fair comparison, both the classical and LLM-based Boids implementations were executed for 5 iterations with 3 boids, repeated for 30 independent trials. Each trial used predefined random seeds so that the initial positions of the boids were generated consistently across both implementations. A relatively small number of iterations was chosen because the number of prompts in the LLM-based setup scales rapidly with each additional iteration and each prompt requires a significant response time. Even with 5 iterations, this configuration resulted in a total of \(3 \text{ boids} \times 3 \text{ prompts} \times 5 \text{ iterations} \times 30 \text{ trials} = 1350\) prompts.

The performance of each system was evaluated using three metrics, with outputs in the range of [0,1]:

\begin{enumerate}
    \item \textbf{Cohesion (C):} Measures how well the boids stay together. A value of 0 indicates complete scattering, while a value of 1 indicates convergence to a single point.
    
    \item \textbf{Separation (S):} Captures the proportion of boid pairs that maintain the minimum required distance. A value of 0 means that all boids collide, while a value of 1 represents perfect spacing.
    
    \item \textbf{Alignment (A):} Quantifies the similarity of velocities among neighboring boids. A value of 0 corresponds to chaotic movement, and a value of 1 indicates complete alignment.
\end{enumerate}

An overall fitness F was calculated as the average of the three metrics:
\begin{equation}
F = \frac{C+S+A}{3}
\end{equation}
Fig.~\ref{fig:boids_classic_llm} presents a performance comparison of classic Boids versus LLM Boids across 30 trials. Classic Boids outperforms LLM Boids in cohesion and separation metrics, resulting in higher F. The larger error bars for LLM Boids in Fig.~\ref{fig:boids_classic} indicate higher performance variability, which is confirmed by the scatter plot of F over trials in Fig.~\ref{fig:boids_llm}, with the solid lines representing the average over all trials. The results will be explained in Sec.~\ref{sec:roleofPE}.
The classical Boids simulation took 3.53 seconds (0.06s per trial) while LLM Boids required 1054.41 seconds (34.98s per trial). For systems requiring real-time operation, the long computation times incurred using LLM Boids are not desirable.

\subsection{Case Study: The ACO Model}
\label{sec:ACOcasestudy}

The ACO model was developed using the methodology outlined in Section~\ref{sec:implementations}. The individual prompts for LLM ACO model are specified in Section~\ref{appendix:boids_prompts}.

For simplicity, the ACO experiment used a fixed network over all iterations -- two candidate paths connecting a common source node to a common destination node using a short and a long path. The interested reader can find all relevant features on our Github repository for reproducibility~\cite{atta2026swarms}. Both the classical and LLM-based ACO implementations were executed for 18 iterations and repeated over 30 trials. For the LLM-based system, three distinct prompts were used, one for path selection, pheromone update, and pheromone evaporation each (see Section~\ref{appendix:boids_prompts}). Predefined random seeds were used for consistent path generation across both implementations. In the LLM-based setup, the number of prompts executed was \(18 \text{ iterations} \times 3 \text{ prompts} \times 30 \text{ trials} = 1620\).

The performance of each implementation was evaluated using four metrics:

\begin{enumerate}
    \item \textbf{Convergence Speed:} Measures how quickly the algorithm consistently favors the optimal path. Convergence occurs when at least 80\% of the most recent selections correspond to the optimal, i.e., short path.

    \item \textbf{Solution Quality:} Measures how strongly the optimal path is reinforced at the end of the algorithm. It is calculated from the final pheromone levels as the ratio of pheromone on the short path to the total pheromone on both paths, with higher values indicating stronger preference for the short path.

    \item \textbf{Learning Efficiency:} Measures how effectively the algorithm transitions from exploration to exploitation during a trial. Ideally, the algorithm explores in the early iterations, transitions in the middle, and primarily exploits in the late iterations. We quantify how closely the model follows this ideal progression across three phases---early, mid, and late---and compute the average score (ranging from 0 to 1) as the learning efficiency. Higher values indicate more effective learning.
    
    \item \textbf{Learning Stability:} Measures how steadily the algorithm progresses towards the optimal. It is based on changes in the pheromone ratio, with higher values indicating more consistent learning and less fluctuations.
\end{enumerate}

Fig.~\ref{fig:aco_performance} presents the comparison of Classic ACO versus LLM ACO for the four metrics. LLM ACO demonstrates superior performance for all metrics with smaller error bars indicating lower trial-to-trial variability and more consistent behavior.

Fig.~\ref{fig:aco_phase} examines the progression from early phase through mid phase to late phase, revealing the learning trajectory of each approach. For 18-iteration runs, LLM ACO exhibits better progression (52.2\% $\rightarrow$ 86.7\% $\rightarrow$ 100.0\% optimal path selection) compared to Classic ACO (61.1\% $\rightarrow$ 71.7\% $\rightarrow$ 72.8\%). By selecting the short path 52\% of the time on average, LLM ACO explores both paths equally in the early phase. Similarly, in the late phase, LLM ACO exploits the optimal path by selecting the short path 100\% of the time. This demonstrates a more effective transition from exploration to exploitation. For classic ACO, the lower performance can be attributed to the low number of iterations. Extending classic ACO beyond 18 iterations (up to 500 iterations) reveals that it approaches near-optimal performance, reaching approximately 98\% optimal path selection by 200 iterations, as illustrated in Fig.~\ref{fig:aco_multirun}. Even with very high number of iterations (500), the total computation time for rule-based ACO is significantly lower than LLM ACO.  For 18-iteration runs, Classic ACO completed all 30 trials in 10.02s (0.334s per trial), while LLM ACO required 1618.48s (53.95s per trial). Extending Classic ACO to 500 iterations incurred negligible additional overhead, completing in 10.25s (0.342s per trial).

Prompt engineering plays a crucial role in the performance of the LLM swarms and will be discussed in detail below.

\subsection{Role of Prompt Engineering}
\label{sec:roleofPE}

The results reported above show that while classic Boids outperforms LLM Boids, LLM ACO attains higher performance than classic ACO. In the Boids implementation, abstract prompts such as ``avoid getting too close'' or ``move slightly'' were used, offering significant freedom in behavior design, but also leading to higher variability in results (see Fig.~\ref{fig:boids_classic}). In contrast, the ACO implementation employed a more structured and context-rich prompt for the path selection agent, which produced more consistent and reliable outcomes. Unlike the classical ACO, which relies on rule-based parameter tuning to balance objectives such as early convergence versus optimal solution quality, the LLM version was explicitly directed to follow specific behaviors. Through prompting, the algorithm was explicitly guided to explore in the early phase, transition in the middle phase, and exploit in the late phase, ensuring optimal performance across all phases regardless of the total number of iterations. The prompt engineering in the two case studies highlights the trade-off between flexibility and control---one can either use open-ended prompts and accept variable results, or invest effort in prompt engineering to achieve a bounded performance.

We further studied the impact of prompt engineering by using different prompts for the same case study. The initial prompt for the path selection agent in LLM ACO (available on our GitHub repository) resulted in much lower learning efficiency compared to the classical ACO, performing 0\% exploration during the trial. The final path selection agent prompt (shown in Section~\ref{appendix:boids_prompts}) achieved the desired efficiency (refer to the discussion on Fig.~\ref{fig:aco_phase} in Sec.~\ref{sec:ACOcasestudy}) and differs substantially from the initial version. Unlike the initial prompt, which required only basic information about the two paths such as path length and pheromones, the final prompt incorporates additional contextual details about the system’s state during each trial, including the current phase and iteration number, enabling more effective decision-making.

\subsection{LLM-powered Swarms: A new frontier or a fad?}

Finally, we examine the term ``swarm'' as adopted within the OAS framework.  In the OAS framework, traditional swarm principles are reinterpreted through language-driven agents. Classical swarms operate based on fundamental principles, which are \textit{local, decentralized interactions}, \textit{simple rules, scalability}, and \textit{emergence}~\cite{schranz2021swarm}. 

OAS achieves swarming through modular agents that process prompts and generate responses for specific behavioral rules. Each \textit{simple rule} is formulated as a dedicated prompt provided to an individual agent whose output is parsed to guide local behavior. Because OAS agents are stateless between calls, they operate without global awareness, responding only to local inputs---analogous to the \textit{localized interactions} in traditional swarms. \textit{Decentralization} is maintained by allowing each agent to function independently, often corresponding to a specific swarm entity or behavior. Through these distributed, prompt-driven interactions, the OAS agents collectively exhibit \textit{emergent behaviors} comparable to those observed in classical swarm systems. Insights into whether the LLM swarm is also \textit{scalable} can only be gathered through implementation as in this work. The results establish that in the current state, ``swarms'' offered by the OAS framework are far from scalable because of the  high computation cost associated with the design of intelligent swarm agents. Even with improving models, the number of computations required to implement a simple swarm rule is much higher using LLM swarms as compared to traditional swarming algorithms. Implementing local LLMs may require specialized hardware, while reliable network links may be necessary for cloud-based LLMs. This poses additional demands on real-world swarms, for example, robotic swarms which are limited in their computation and payload capacity. In current form, LLM-powered swarms are inefficient in implementing swarming in real-world applications.

\section{Conclusion}
\label{sec:Conclusions}
The comparative study of Boids and ACO highlights a central theme: LLM-based swarms are not yet ready to replace traditional swarms in terms of real-world implementation, but they enrich the design space. Their ability to interpret high-level instructions introduces new flexibility, while rule-based methods remain unmatched in speed and scalability.

A promising direction for future work is the development of hybrid swarm architectures, where LLMs provide high-level strategic reasoning and classical algorithms manage low-level execution, or vice versa. For instance, LLMs could generate adaptive movement strategies while classical algorithms handle fast collision avoidance, or classical models could define global coordination patterns while LLMs manage context-dependent, high-level decision-making. Such approaches could combine the scalability and efficiency of classical swarms with the context-awareness, and adaptive reasoning offered by LLMs. This also underscores the potential role of explainable AI: if models can clarify how specific responses are generated, this knowledge could guide prompt design, reduce trial and error, and allow the human-in-the-loop to refine system behavior more effectively.

This paper shows that the applicability of the LLM-based OAS framework is constrained by high computational costs and hardware requirements, compromising a fundamental swarming principle: scalability. For instance, for the Boids simulation, the LLM-based implementation is approximately 300 times slower than the classical version. This limits scalability, and also hinders deployment, especially in time-sensitive applications. Overall, while current LLM-based swarms face practical limitations, they open new avenues for research, particularly in hybrid designs, prompt engineering, and explainable AI for LLM-based swarms.

\section*{Acknowledgment}

The authors would like to thank Oleksandr Chepizhko for providing a comprehensive review. The work is funded by the European Union, project MYRTUS, by grant No. 101135183. Views and opinions expressed are however those of the author(s) only and do not necessarily reflect those of the European Union. Neither the European Union nor the granting authority can be held responsible for them.


\bibliographystyle{IEEEtran}
\bibliography{references}

\begin{IEEEbiographynophoto}{Muhammad Atta Ur Rahman} is a researcher at Lakeside Labs at Klagenfurt, 9020, Austria. His research interests include artificial intelligence, robotics, and multi-agent systems, with a focus on developing intelligent, collaborative agents for autonomous decision-making and control. Rahman received his Masters degree in Autonomous Systems and Robotics from University of Klagenfurt.
\end{IEEEbiographynophoto}

\begin{IEEEbiographynophoto}{Dr. Melanie Schranz} is senior researcher at Lakeside Labs, Klagenfurt, 9020, Austria. Her research interests include engineering swarm intelligence for cyber-physical systems, the abstract simulation and integration to real-world use cases of these algorithms. Schranz is a member of Discover US and Nexus Forum, European initiatives to foster distributed computing.
\end{IEEEbiographynophoto}

\begin{IEEEbiographynophoto}{Dr. Samira Hayat} is a senior researcher at Lakeside Labs, Klagenfurt, 9020, Austria. Her research interests include drone swarm designs, with the focus on the role of communication in swarm coordination, and innovative use case development. She was nominated as``Rising star in Computer Networking and Communications'' by IEEE N2Women in 2022.
\end{IEEEbiographynophoto}
\vfill

\end{document}